# Fault diagnosis for three-phase PWM rectifier based on deep feedforward network with transient synthetic features

Lei Kou [a], Chuang Liu [a,*], Guo-wei Cai [a], Zhe Zhang [b], Jia-ning Zhou [a], Xue-mei Wang [a]

[a] *School of Electrical Engineering, Northeast Electric Power University, Jilin 132012, China*
[b] *Department of Electrical Engineering, Technical University of Denmark, 2800 Kgs. Lyngby, Denmark*

## ABSTRACT

Three-phase PWM rectifiers are adopted extensively in industry because of their excellent properties and potential advantages. However, while the IGBT has an open-circuit fault, the system does not crash suddenly, the performance will be reduced for instance voltages fluctuation and current harmonics. A fault diagnosis method based on deep feedforward network with transient synthetic features is proposed to reduce the dependence on the fault mathematical models in this paper, which mainly uses the transient phase current to train the deep feedforward network classifier. Firstly, the features of fault phase current are analyzed in this paper. Secondly, the historical fault data after feature synthesis is employed to train the deep feedforward network classifier, and the average fault diagnosis accuracy can reach 97.85% for transient synthetic fault data, the classifier trained by the transient synthetic features obtained more than 1% gain in performance compared with original transient features. Finally, the online fault diagnosis experiments show that the method can accurately locate the fault IGBTs, and the final diagnosis result is determined by multiple groups results, which has the ability to increase the accuracy and reliability of the diagnosis results.

## 1. Introduction

Three-phase PWM voltage-source rectifier(VSR) has been widely adopted in the fields of high-speed rails, aerospace and renewable energy system because of its excellent performance and potential advantages such as simple circuit structure, good controllability and so forth [1–3]. Although there are several ways to strengthen the reliability of the system, fault is inevitable. It is necessary for early fault detection to avoid catastrophic damage, meanwhile it is also extremely important to locate fault position and reduce maintenance time quickly [4].

The features of open-circuit faults (OCFs) and short-circuit faults (SCFs) in IGBTs are studied in [5] and [6], where SCFs also cause over-current in most cases, therefore, the SCFs protection are usually completed by standard hardware circuits [7]. However, OCFs are usually caused by the faults of power-driven circuits or power devices, as OCFs will not cause the system to crash immediately, may not be detected for a long time, so it is easy to lead the secondary fault to other devices and result in higher maintenance costs [8,9]. A method of turning SCFs into OCFs by using fast fuse is introduced in [10], which can reduce the damage of SCFs. Since the OCFs may not be detected by the standard protection circuit, it is necessary to study the early fault features of OCFs in power devices to improve the reliability of the power electronic system [11].

In general, fault diagnosis methods mainly include model-based methods and data-driven methods [12–14]. The detailed analysis about fault mechanism of 3-phase PWM VSR is given in [15], when OCFs happen in one or more transistors of the rectifier, the diode will continue to work as the rectifier element, and the system will not collapse immediately, but the performance will be reduced, such as output voltage fluctuations and current harmonics. The AC side current of the PWM VSR is not always zero while the OCFs happen in IGBT, but is a great extent to approach the sine negative half cycle of the non fault condition, which is not the same as the ideal mathematical model. On the basis of current Luenberger observer model and adaptive thresholds, an OCFs diagnosis method was presented in [16] for PMSG drives. In order to reduce the dependence on the fault mathematical models, it has been a trend to assist fault diagnosis with machine learning algorithm, and machine learning methods can distinguish different fault mechanisms in a complex system, which does not need too much dependence on mathematical models, only uses historical data [17]. And it is possible to use simulation tools to simulate approximate historical data of different fault states [18,19]. With the development of information and computer technology, it is no longer difficult to collect large quantities of fault data, and data-driven methods have also gained a lot of attention [20–26]. An AI-based fault diagnosis and reconfiguration method is presented in [22] for multilevel inverter drive (MLID), the phase voltages after dimension reduction with PCA (Principal Component Analysis) were employed as diagnosis variables, and the diagnosis accuracy can be reach higher than 95%. Based on artificial neural network (ANN) method, an OCFs diagnosis method was proposed in [23] for any switch in the microgrid inverter. Based on main fault component analysis method, an OCFs diagnosis algorithm for switches of microgrid inverter was presented in [24]. A knowledge-based method was proposed in [25] to normalized the phase currents, and then random forests(RFs) algorithm was used to train the data-driven fault diagnosis classifier, but the method is only suitable for 3-phase system because it depends on some knowledge. Based on the support vector machine (SVM) method, a data-driven faults detection and diagnosis method was proposed in [26] for electrical drives of trains. And [24] and [26] were verified based a dSPACE digital controller.

Most scholars mainly study inverter fault, but few researches are about rectifier circuit fault, meanwhile it is difficult to extend the application of mathematical models [27,28]. And what is more, the transient values related to OCFs are usually larger than those of steady state [29]. Therefore, the features of OCFs in IGBTs of 3-phase PWM VSRs are deeply studied, and then a novel OCFs diagnosis method is presented based on deep feedforward network with transient synthetic features. According to a research on the reliability of power electronic converters in the industrial field, IGBT has become one of the most widely used power semiconductor devices [30]. The probability of simultaneous fault in multiple IGBTs is also relatively low [31,32]. Therefore, the case of single or double transistor faults is the focus of research. The main contribution of the subject is the innovative interdisciplinary application. The deep feedforward network (DFN) algorithm has been widely used for fault diagnosis, but most of them are used for fault diagnosis and detection in power system, and few scholars have studied its application for 3-phase PWM VSRs. In addition, some fault diagnosis methods rely on the fault models of power electronic converters, such as [16]. And some fault diagnosis methods do not take into account the cost of the operating environment, such as [24] and [26]. Compared with [25], the transient synthetic features method of this paper can produce the new features and add the dimension of features, which has greater significance to improve the fault diagnosis accuracy. And the proposed method has stronger general ability, which can be applied to most of power electronic systems. Each method has its own research background, and they have achieved good results, which give us a lot of inspiration and experience. Therefore, based on deep feedforward network, a novel OCFs diagnosis method is presented for 3-phase PWM VSRs, and the synthetic features method is adopted to improve the diagnosis accuracy. And the contributions of the study are listed as follows:

(1) Only the currents used in closed-loop control are selected as the diagnosis variables, and additional instrumentation is not needed in this study. The deep feedforward network is trained by the historical fault data, which is a data-driven method. It can reduce the dependence on the fault mathematical models of power electronics circuit, because the fault mathematical model is difficult to establish.

(2) The transient synthetic fault features method is adopted to improve the classification ability of the fault diagnosis classifier.

(3) Since most AI(artificial intelligence) algorithms can only run on computer, the time scale is considered to reduce the data

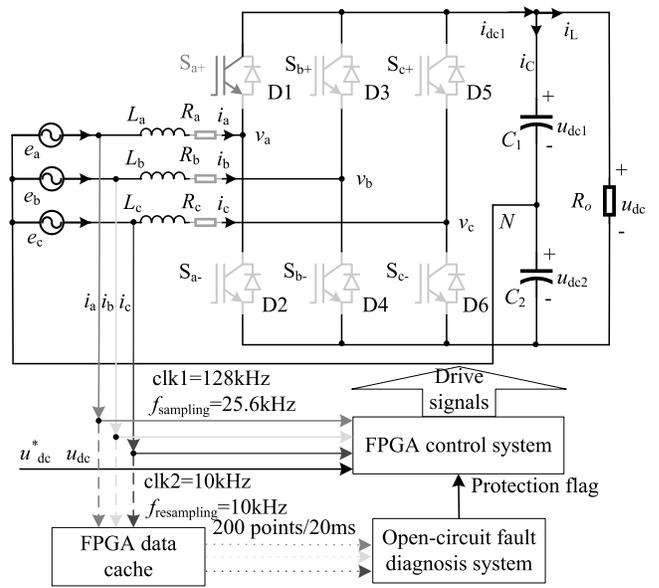

**Fig. 1.** OCFs diagnosis system for 3-phase PWM VSRs.

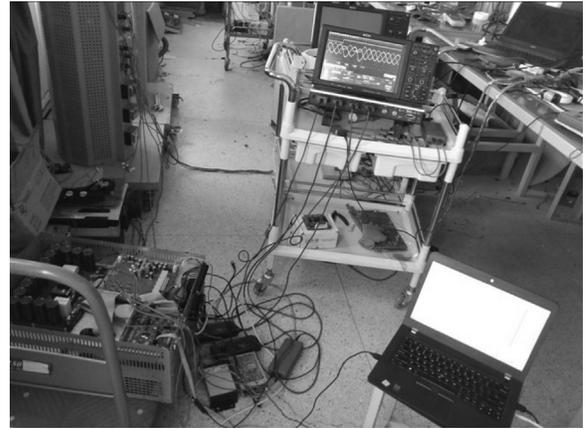

**Fig. 2.** Experimental system.

transmission rate. And the online fault diagnosis system based on DFN with transient synthetic features can obtain 200 diagnosis results per 20 ms. The final fault locations are achieved by using 200 diagnosis results, the final result can be improved by this way.

The rest of the paper is arranged as follows: In Section 2, the fault data of phase current and output voltage in the 3-phase PWM VSR are collected when OCFs occur in IGBTs. In Section 3, firstly, the fault features are analyzed and extracted, and then, an OCFs diagnosis method based on deep feedforward network with transient synthetic features is proposed, meanwhile various OCFs diagnosis classifiers are compared and analyzed. In Section 4, the online OCFs diagnosis and locations experiments are completed, the monitoring and fault diagnosis system is developed, and the proposed method is validated by test. Finally, the conclusion is presented in Section 5.

## 2. OCFs data acquisition based on actual experiments

In order to be consistent with the actual situation as far as possible, the fault samples of this paper is directly from experiments, which is obtained through the actual fault simulation experiments. Because of the reversibility between rectifier and inverter of the grid-side converter, a 3-phase PWM VSR is designed to work in the rectification state to study the problems of OCFs in IGBTs. Fig. 1 shows the process of OCFs diagnosis system for 3-phase PWM VSRs, where the OCF happens in $S_{a+}$, and the D1 is still working normally. In this paper, the Proportional-Resonant (PR) controller is used to complete the control of 3-phase PWM VSRs, which can refer to [33].

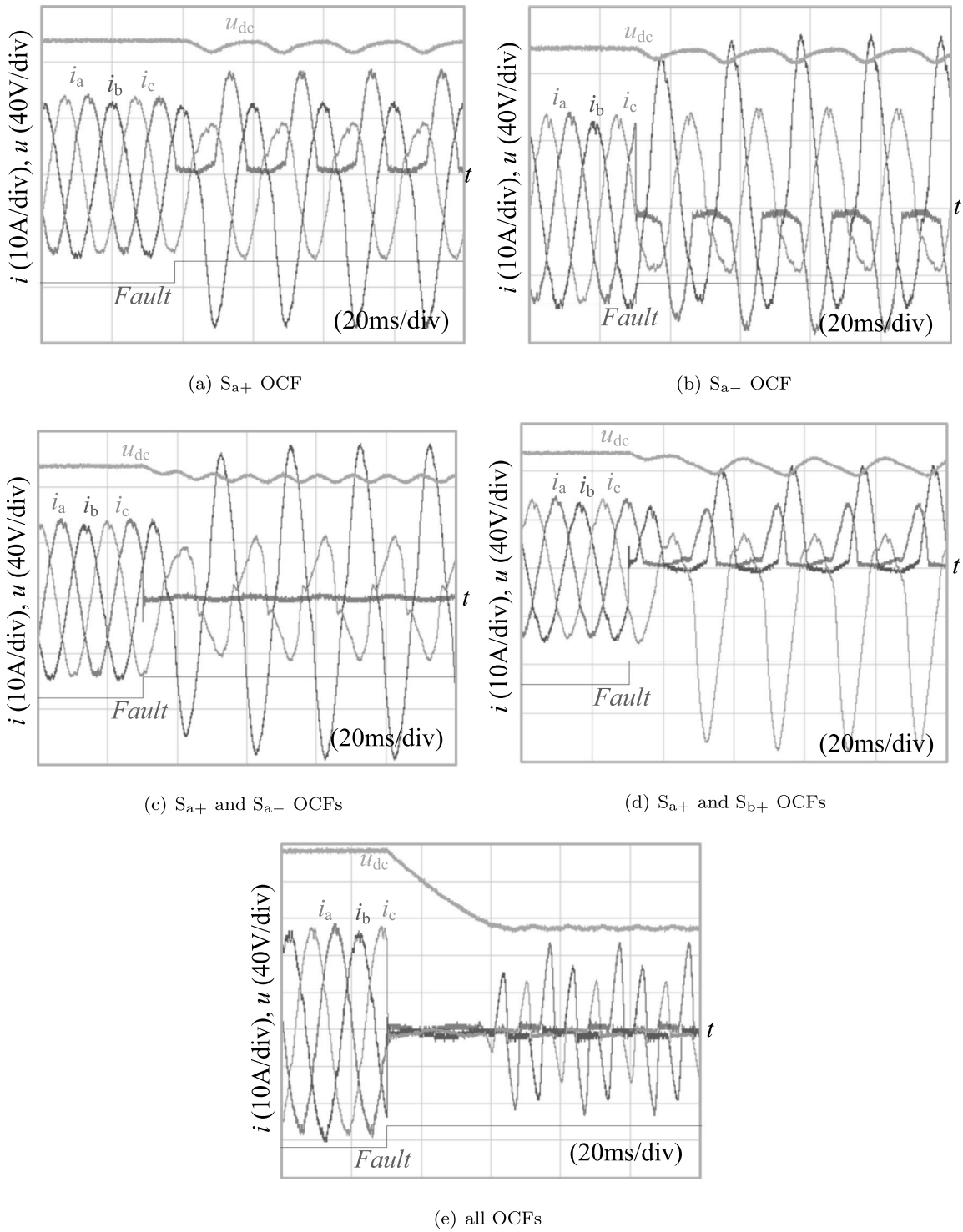

Fig. 3. Some OCFs waveform.

The parameters of experimental system are shown in Table 1, and Fig. 2 is the experimental system of 3-phase PWM VSRs. For safety consideration, the control software is used to block control signals of the IGBTs to simulate the OCFs happen in IGBT, and the sampling frequency of the oscilloscope is 25MHz. Some OCFs phase currents and OCFs output voltages waveforms are shown in Fig. 3, and the data are exported for training deep feedforward network. According to the fault waveform, the phase will soon exceed twice the normal value when the fault occurs, which will lead secondary fault to the whole system. And the fault waveform and fault analysis will be discussed in detail in Section 3.

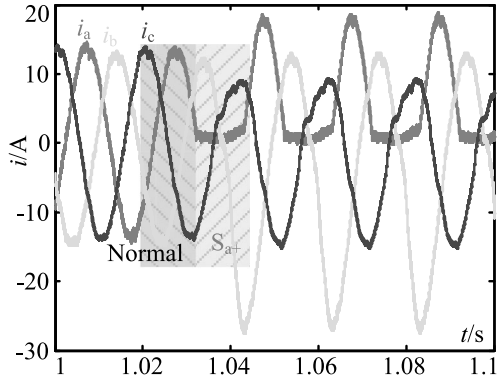

(a) $S_{a+}$ OCF

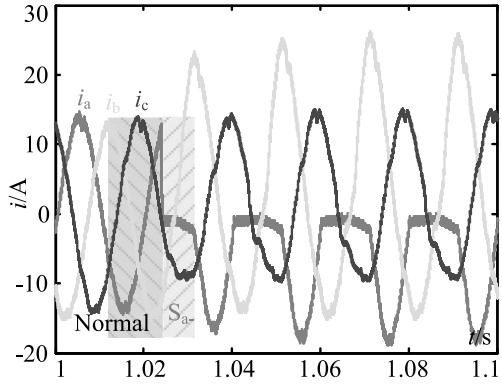

(b) $S_{a-}$ OCF

**Fig. 4.** Waveforms of OCF in single IGBT.

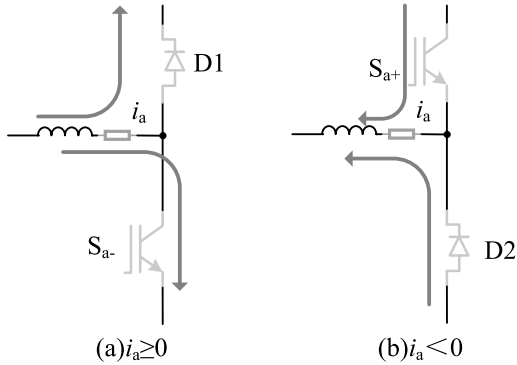

(a) $i_a \geq 0$      (b) $i_a < 0$

**Fig. 5.** The conducting route under different directions of $i_a$.

## 3. Training and evaluation of OCFs diagnosis classifier based on deep feedforward network

In this section, the fault features are analyzed and extracted, and then, the fault diagnosis classifiers based on DFN, RFs and SVM algorithms are compared and evaluated. Meanwhile the accuracies of fault diagnosis methods are improved by using synthetic features.

### 3.1. Features of OCFs in IGBTs of 3-phase PWM VSRs

The power electronic devices are the core components of 3-phase PWM VSRs, therefore, the corresponding fault feature analysis is one of the important bases for fault diagnosis in the power electronic circuit [34]. Fig. 4 shows the phase current while the OCF happens in a single IGBT, where $i_a$, $i_b$, $i_c$ are phase currents respectively. The phase currents show different features when OCFs occur in different IGBTs. According to Fig. 4, when the OCFs happen in the upper IGBTs, the current waveform of the lower half period of the corresponding phase current is affected, and the lower arm affects the waveform of the upper half period.

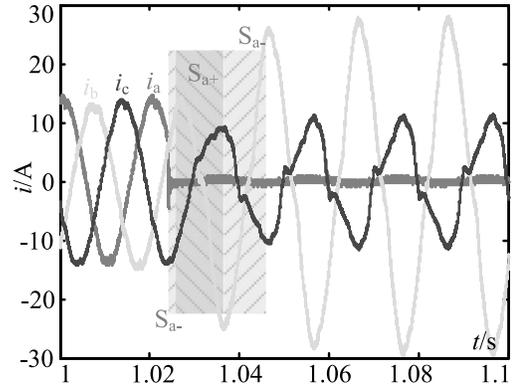

(a) $S_{a+}$ and $S_{a-}$ OCFs

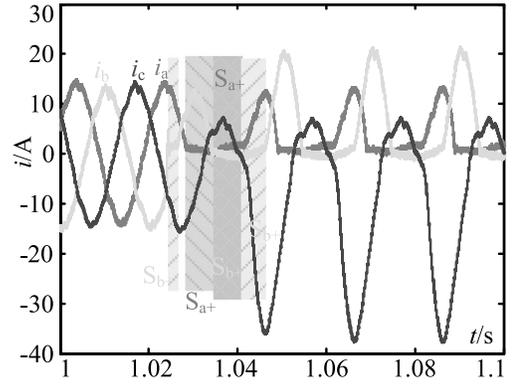

(b) $S_{a+}$ and $S_{b+}$ OCFs

**Fig. 6.** Some IGBTs OCFs at the same time.

**Table 1**
Parameters of 3-phase PWM VSRs.

| Parameter | Value |
| --- | --- |
| Input voltage | 40 V and 50 Hz |
| Filter inductance | 500 μH |
| Output voltage | 100 V |
| DC-link capacitances | 7000 μF |
| Sampling frequency | 25.6 kHz |
| IGBT | FS1501212ET4 |
| Switching frequency | 12.8 kHz |
| Load | 16 Ω |

Taking the current $i_a$ as an example, the reason for the above phenomenon is that the current $i_a$ in the PWM rectifier consists of the following channels (as shown in Fig. 5). (see Table 2).

Fig. 6 shows the phase currents features while OCFs happen in two IGBTs simultaneously, but the fault features will not always be displayed at the same time. Where Fig. 6(a) shows the OCFs occur in $S_{a+}$ and $S_{a-}$ simultaneously, the fault features of $S_{a-}$ fault are displayed first, followed by the fault features of $S_{a+}$, and so on. Fig. 6(b) shows the OCFs happen in $S_{a+}$ and $S_{b+}$ simultaneously, the order of fault features is $(S_{b+}) \rightarrow (\text{normal}) \rightarrow (S_{a+}) \rightarrow (S_{a+}$ and $S_{b+}) \rightarrow (S_{b+})$.

**Table 2**
The reason of fault features.

| Conditions | Current direction | Influence |
|---|---|---|
| $i_a \geq 0$, $S_{a+}$ turn-on and $S_{a-}$ turn-off | $i_a$ flows through equipped diode D1 | $S_{a+}$ and $S_{a-}$ no effect |
| $i_a \geq 0$, $S_{a+}$ turn-off and $S_{a-}$ turn-on | $i_a$ flows through IGBT $S_{a-}$ | $S_{a+}$ no effect, $S_{a-}$ has effect |
| $i_a < 0$, $S_{a+}$ turn-on and $S_{a-}$ turn-off | $i_a$ flows through IGBT $S_{a+}$ | $S_{a+}$ has effect, $S_{a-}$ no effect |
| $i_a < 0$, $S_{a+}$ turn-off and $S_{a-}$ turn-on | $i_a$ flows through equipped diode D2 | $S_{a+}$ and $S_{a-}$ no effect |

**Table 3**
Codes and labels of OCFs.

| Fault IGBT | Fault code | Fault label |
|---|---|---|
| Normal | F0 | 0 |
| $S_{a+}$ | F1 | 1 |
| $S_{a-}$ | F2 | 2 |
| $S_{b+}$ | F3 | 3 |
| $S_{b-}$ | F4 | 4 |
| $S_{c+}$ | F5 | 5 |
| $S_{c-}$ | F6 | 6 |
| $S_{a+}$, $S_{b+}$ | F7 | 7 |

Fig. 7 shows the phase current waveform while OCFs happen in multiple IGBTs. According to Figs. 4, 6 and 7, while OCFs happen in two or more IGBTs at the same time, the features of OCF in one IGBT will be shown at first, and then the superposition features, finally the features of OCF in another IGBT shown.

Fig. 8 shows the output voltage $u_{dc}$ while OCFs happen in some IGBTs, and it is obvious that the rectifier will run in uncontrolled mode only when OCFs happen in all IGBTs. The output voltage can still work in controlled mode when the OCFs occur in not all IGBT. Therefore, the rectifier can still operate with fault when a fault occurs, which will bring security risks to the whole system.

According to the above research, the fault features are not usually displayed simultaneously in the early stage when OCFs happened in two IGBTs at the same time. Moreover the most important and the worst issue is that the output voltage will be normal as long as OCFs occur in not all IGBT at a short time, so it will not cause system protection, and the fault is not discovered. However, the long-term operation of faults will accelerate the damage of other devices and leave a hidden danger to the power electronic system. Thus, it is very important to monitor and locate the OCFs in time.

### 3.2. Training of OCFs diagnosis classifier

The OCFs diagnosis classifiers trained by the time series, original transient features and transient synthetic features are compared in this paper, the additional synthetic features can be used to enhance the classification ability of the OCFs diagnosis classifier, and the synthetic feature can be achieved by arithmetic operations such as addition, subtraction, multiplication, division, the synthetic features can be seen as hidden features extracted by the neural networks. Addition and subtraction are linear operations, and division has the disadvantage when the divisor is 0. Therefore, multiplication is adopted to synthesize features.

The transient data of three phase current under different fault conditions are used to train the OCFs diagnosis classifier. It is found that some fault features can be displayed simultaneously while two or more IGBTs such as OCFs occur in $S_{a+}$ and $S_{b+}$. But some fault features will not be displayed at the same time (for example, $S_{a+}$ and $S_{a-}$). Therefore, the study mainly focuses on the OCF in single IGBT, and takes the OCFs of $S_{a+}$ and $S_{b+}$ as examples for OCFs diagnosis. To facilitate training, the faults are coded and tagged (as shown in Table 3).

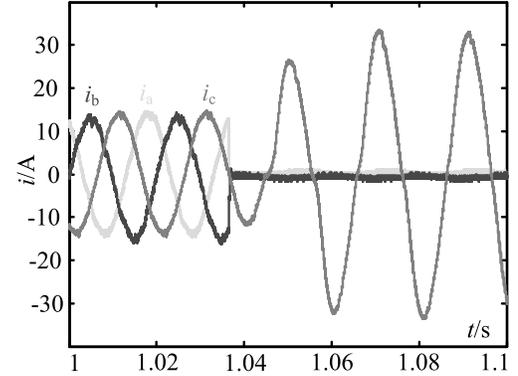

(a) $S_{a+}$, $S_{a-}$, $S_{b+}$ and $S_{b-}$ OCFs

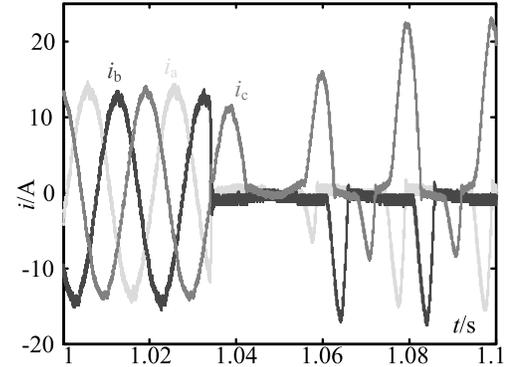

(b) $S_{a+}$, $S_{a-}$, $S_{b+}$, $S_{b-}$ and $S_{c+}$ OCFs

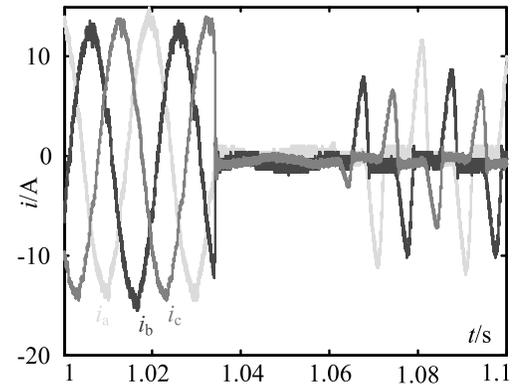

(c) all IGBTs OCFs

**Fig. 7.** Waveforms of multiple IGBTs OCFs.

When implementing the fault diagnosis classifier, normalization should first be taken into account to normalize the fault samples to $[x'_{min}, x'_{max}]$, where $x'_{min} = -1$, $x'_{max} = 1$. Thus, it can narrow the numerical difference among the fault samples, reduce the training error caused by the large numerical difference of samples values, and raise the accuracy of the OCFs diagnosis classifier. Where the expression of the normalization process is:

$$x' = \begin{cases} \frac{(x'_{max} - x'_{min})(x - x_{min})}{(x_{max} - x_{min})} + x'_{min}, x_{max} \neq x_{min} \\ x'_{min}, x_{max} = x_{min} \end{cases} \quad (1)$$

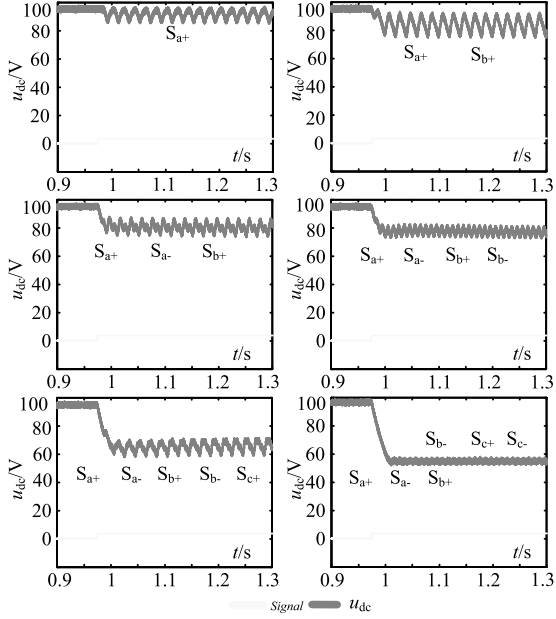

**Fig. 8.** Output voltage under OCFs states.

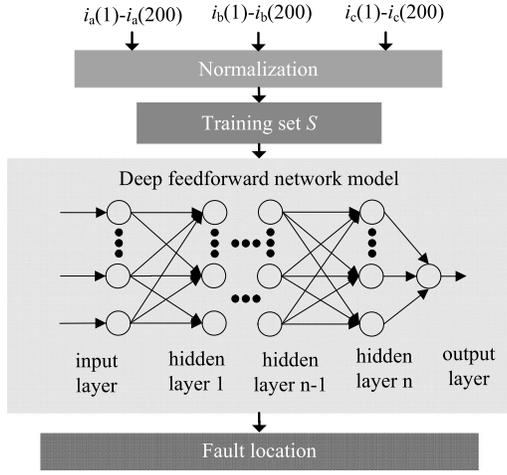

**Fig. 9.** Traditional structure of deep feedforward network with time series.

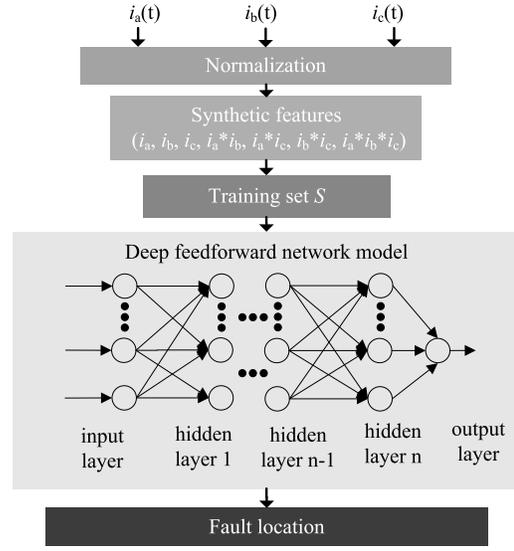

**Fig. 10.** Structure of deep feedforward network with transient synthetic features.

Where $x$ is the fault sample data, $x_{min}$ and $x_{max}$ are the minimum and maximum value of the samples in the same group, respectively, $x'$ is the sample after normalization.

Fig. 9 shows the traditional structure of deep feedforward network with time series, and Fig. 10 shows the structure of deep feedforward network with transient synthetic features. As shown in Figs. 9 and 10, the biggest difference is that the traditional input features are time series, and transient synthetic features are employed as the input of the proposed method. Transient features require fewer samples to be measured directly than time series, and the synthetic features are adopted to increase the number of features.

The setting of parameters and the selection of activation function can be referred to [35] and [36], and it is also chosen by the user's experience. The parameters of deep feedforward network are as follows: The training samples are taken from the normal state and F0-F7 states, the input samples are ($i_a$, $i_b$, $i_c$) or the synthetic features ($i_a$, $i_b$, $i_c$, $i_a*i_b$, $i_a*i_c$, $i_b*i_c$, $i_a*i_b*i_c$), and the output is the label value of each sample. The strong nonlinear modeling ability of the deep feedforward network is benefit from the combination of various activation functions, in which the function Tansig and function purelin are employed in the hidden and output layer, respectively, and the number of neurons in each layer are [16, 16, 16, 16, 16, 16, 16, 16, 16, 16]. The training accuracy is set to 0.0001, and the learning efficiency is set to 0.01. And the detailed training process of SVM algorithm can refer to [37], the detailed training process of RFs algorithm can refer to [14].

### 3.3. Evaluation of fault diagnosis classifiers

The fault diagnosis classifiers are achieved by training the deep feedforward network with time series, ($i_a$, $i_b$, $i_c$) and the transient synthetic features ($i_a$, $i_b$, $i_c$, $i_a*i_b$, $i_a*i_c$, $i_b*i_c$, $i_a*i_b*i_c$), and the fault classifier can be replaced by a black box function $f(x)$, and the diagnosis function can be expressed as

$$y = \begin{cases} round(f(x)), -0.5 < f(x) < 7.5 \\ error, otherwise \end{cases} \quad (2)$$

Where $x$ is time series, ($i_a$, $i_b$, $i_c$) or the transient synthetic features, $y$ is the label of the final output result, and the output results of $f(x)$ are approximate to the fault labels $y$, the round function provides a rounded-data. As shown in Table 4 is the diagnosis result of some samples.

Table 5 shows the performance of various methods. According to Table 5, the classification accuracy of DFN with synthetic features is the highest, and the classification accuracy of SVM is lower than that of DFN or RFs. The training time of RFs is the shortest, and the training time of SVM with time series is the longest. For power electronic fault diagnosis, the diagnosis accuracy is more important than the training speed. And when the feature dimension is high and the number of samples is sufficient, the DFN can get higher diagnosis accuracy than RFs and

**Table 4**
Some samples and diagnosis results.

| Fault IGBT | $i_a$ | $i_b$ | $i_c$ | $f(x)$ | $y$ |
|---|---|---|---|---|---|
| F0 | −2.98 | −11.62 | 14.28 | −0.07 | 0 |
| F1 | 0.33 | 13.28 | −11.95 | 1.02 | 1 |
| F2 | −1.66 | −14.94 | 12.95 | 1.99 | 2 |
| F3 | −11.95 | 0.66 | 13.95 | 2.96 | 3 |
| F4 | 11.62 | −1.66 | −13.28 | 4.01 | 4 |
| F5 | 13.28 | −12.95 | 1.33 | 5.01 | 5 |
| F6 | −14.28 | 11.62 | −0.66 | 5.99 | 6 |
| F7 | 0.99 | −0.66 | 2.99 | 6.97 | 7 |

**Table 5**
Performance of classifiers.

| Methods | Average accuracy | Training time |
|---|---|---|
| DFN + time series | 0.9753 | 85.6 min |
| DFN + $i_a$, $i_b$, $i_c$ | 0.9666 | 30.5 min |
| DFN + synthetic features | 0.9785 | 26.7 min |
| RFs + time series | 0.9678 | 10.3 min |
| RFs + $i_a$, $i_b$, $i_c$ | 0.9682 | 4.7 min |
| RFs + synthetic features | 0.9702 | 6.2 min |
| SVM + time series | 0.9487 | 110.5 min |
| SVM + $i_a$, $i_b$, $i_c$ | 0.9465 | 42.3 min |
| SVM + synthetic features | 0.9523 | 36.6 min |

**Table 6**
Diagnosis results' probability of the classifier with time series.

| Fault | F0 | F1 | F2 | F3 | F4 | F5 | F6 | F7 |
|---|---|---|---|---|---|---|---|---|
| F0 | 0.9571 | 0.0264 | 0.0057 | 0.0037 | 0.0033 | 0.0021 | 0.0017 | 0 |
| F1 | 0.0049 | 0.9431 | 0.0125 | 0.0196 | 0.0079 | 0.0063 | 0.0035 | 0.0022 |
| F2 | 0.0012 | 0.0256 | 0.9698 | 0.0013 | 0.0004 | 0.0005 | 0.0012 | 0 |
| F3 | 0.0009 | 0.0014 | 0.0016 | 0.9762 | 0.0048 | 0.0042 | 0.0051 | 0.0058 |
| F4 | 0.0005 | 0.0013 | 0.0024 | 0.0038 | 0.9874 | 0.0021 | 0.0010 | 0.0015 |
| F5 | 0.0005 | 0.0021 | 0.0064 | 0.0120 | 0.0102 | 0.9654 | 0.0011 | 0.0023 |
| F6 | 0.0008 | 0.0013 | 0.0035 | 0.0013 | 0.0018 | 0.0017 | 0.9874 | 0.0022 |
| F7 | 0 | 0.0041 | 0.0031 | 0.0142 | 0.0048 | 0.0031 | 0.0273 | 0.9434 |

**Table 7**
Diagnosis results' probability of the classifier with ($i_a$, $i_b$, $i_c$).

| Fault | F0 | F1 | F2 | F3 | F4 | F5 | F6 | F7 |
|---|---|---|---|---|---|---|---|---|
| F0 | 0.9557 | 0.0311 | 0.0058 | 0.0045 | 0.0013 | 0.0004 | 0.0012 | 0 |
| F1 | 0.0100 | 0.9341 | 0.0132 | 0.0243 | 0.0085 | 0.0054 | 0.0028 | 0.0017 |
| F2 | 0.0019 | 0.0308 | 0.9663 | 0.0010 | 0 | 0 | 0 | 0 |
| F3 | 0 | 0.0002 | 0.0005 | 0.9802 | 0.0056 | 0.0032 | 0.0037 | 0.0066 |
| F4 | 0.0001 | 0 | 0.0001 | 0.0043 | 0.9954 | 0.0001 | 0 | 0 |
| F5 | 0.0003 | 0.0014 | 0.0078 | 0.0107 | 0.0093 | 0.9704 | 0.0001 | 0 |
| F6 | 0.0006 | 0 | 0 | 0.0013 | 0.0008 | 0.0017 | 0.9956 | 0 |
| F7 | 0 | 0.0014 | 0.0045 | 0.0207 | 0.0098 | 0.0151 | 0.0273 | 0.9212 |

**Table 8**
Diagnosis results' probability of the classifier with transient synthetic features.

| Fault | F0 | F1 | F2 | F3 | F4 | F5 | F6 | F7 |
|---|---|---|---|---|---|---|---|---|
| F0 | 0.9842 | 0.0017 | 0.0084 | 0.0031 | 0.0013 | 0.0004 | 0.0009 | 0 |
| F1 | 0.0067 | 0.9582 | 0.0195 | 0.0034 | 0.0037 | 0.0046 | 0.0021 | 0.0018 |
| F2 | 0.0010 | 0.0290 | 0.9695 | 0.0005 | 0.0000 | 0 | 0 | 0 |
| F3 | 0.0000 | 0.0000 | 0.0004 | 0.9808 | 0.0041 | 0.0042 | 0.0007 | 0.0098 |
| F4 | 0.0002 | 0.0000 | 0.0002 | 0.0009 | 0.9987 | 0.0000 | 0 | 0 |
| F5 | 0.0007 | 0.0008 | 0.0107 | 0.0067 | 0.0081 | 0.9722 | 0.0008 | 0 |
| F6 | 0.0006 | 0.0001 | 0.0008 | 0.0000 | 0.0000 | 0.0007 | 0.9978 | 0 |
| F7 | 0 | 0.0016 | 0.0080 | 0.0123 | 0.0162 | 0.0131 | 0.0057 | 0.9431 |

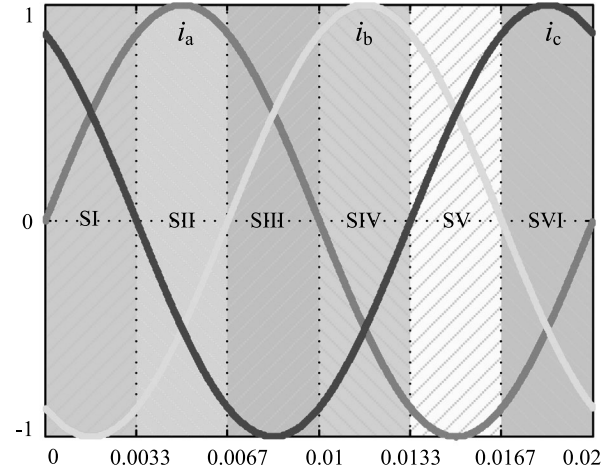

**Fig. 11.** Distribution of IGBT OCFs.

SVM. Therefore, the performance of DFN algorithm with synthetic features is the best. Compared with original features, the DFN classifier trained by the synthetic features obtained more than 1% gain in performance.

Tables 6–8 show the diagnosis results' probability of the DFN classifiers which trained by time series, ($i_a$, $i_b$, $i_c$) and transient synthetic features respectively, and a total of 2.05 million samples participated in the test.

According to Tables 6–8, the probability of classification is high, and the probability of misdiagnosis is generally low, which fully proved that the deep feedforward network is suitable for OCFs diagnosis of power switches. It is obvious that the training dataset with synthetic features can boost the classifier's performance up. Comparing the three results, it can be seen that the classifier trained by the transient synthetic features is more precise, and the diagnosis accuracy is relatively balanced for F0-F7 states. The method with transient synthetic features offers extremely promising discrimination accuracy even with training dataset of less than 5% of total samples (0.1025 million out of 2.05 million total samples) compared with test/validation dataset, which gives more than 97% average accuracy for different faults.

## 4. Online OCFs diagnosis and locations experiments

According to the analysis of OCFs in IGBTs, it is known that the waveform of the lower half period of the phase current will be affected when the OCFs happen in the upper IGBTs, and the upper half period currents will be affected by the lower IGBTs. According to the above rules, the time distribution diagram of fault waveform in ideal state can be expressed in Fig. 11. Any combination of ($S_{a-}$, $S_{b+}$, $S_{c+}$) faults can be diagnosed from the SI region in Fig. 11. The SII region corresponds to the combination of ($S_{a-}$, $S_{b-}$, $S_{c-}$). Similarly, the SIII, SIV, SV and SVI regions also have the corresponding combinations. It cannot detect the fault in $S_{a+}$ if fault occurs at SI region by transient fault features. Therefore, a period of fault diagnosis results is taken into account as a reference to determine the final diagnosis result, and then OCFs in multiple IGBTs can be diagnosed and located.

The experimental workbench of the 3-phase PWM VSR system as shown in Fig. 2. As shown in Fig. 1, the time scale is considered because most AI(artificial intelligence) algorithms can only run on computer, where the FPGA controller can send 200 data points per 20 ms to the computer through the network, and the fault diagnosis system based on deep feedforward with transient synthetic features can obtain 200 diagnosis results per 20 ms. The final fault location and diagnosis of multiple faults are achieved by using 200 diagnosis results, and meanwhile the final result is enhanced by the way.

Fig. 12 shows the OCFs diagnosis process when OCFs occur in $S_{a+}$ and ($S_{a+}$ and $S_{b+}$), and Fig. 13 shows the OCFs diagnosis

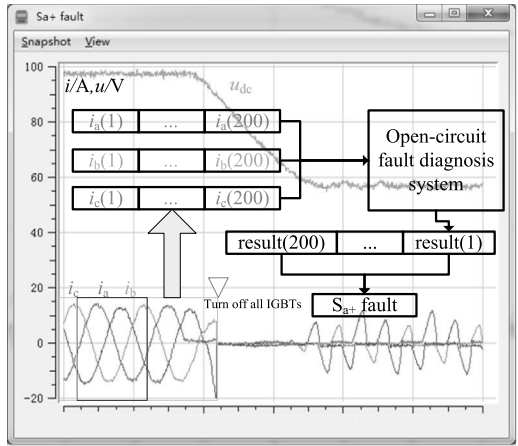

(a) $S_{a+}$ OCF

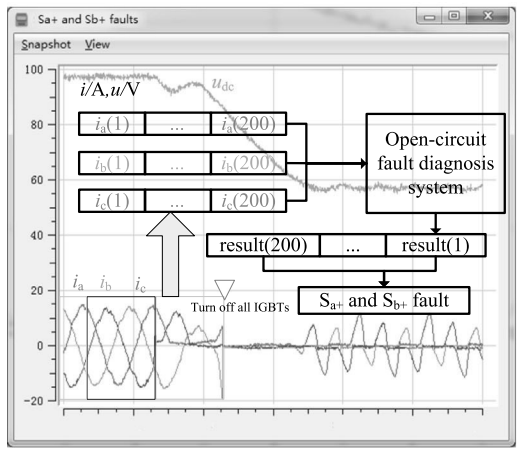

(b) $S_{a+}$ and $S_{b+}$ OCFs

**Fig. 12.** Fault diagnosis process.

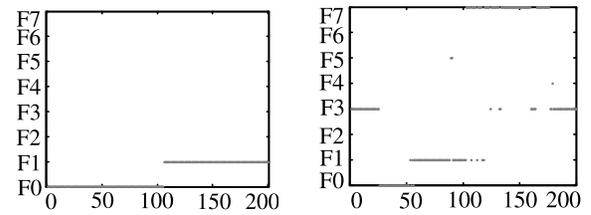

(a) $S_{a+}$ OCF diagnosis results

(b) $S_{a+}$ and $S_{b+}$ OCFs diagnosis results

**Fig. 13.** F1 and F7 OCFs diagnosis results.

results. As shown in Fig. 13(a), when an OCF happens in a single IGBT, the distribution of OCFs diagnosis results approximates to the normal state and the fault state accounts for half of each, and consistent with previous fault feature analysis. As shown in Fig. 13(b), while the occurrence of the OCFs in $S_{a+}$ and $S_{b+}$ at the same time. At the beginning, the faults features are not the OCFs in two IGBTs. The order of diagnosis is F3-F0-F1-F7-F3. By this way, both OCFs in $S_{a+}$ and $S_{b+}$ can be diagnosed. The traditional time series is that when the 200 sets of samples are input into the fault diagnosis system, it can only output a diagnosis result, once the error occurs, the fault location is wrong. However, the transient features can give 200 sets of diagnosis results, multiple groups diagnosis results to determine the final diagnosis result can make the final result more reliable.

## 5. Conclusion

The fault diagnosis method based on deep feedforward network with transient synthetic features is proposed in this paper. The deep feedforward network algorithm, a data-driven method, is trained by the historical fault data, which can reduce the dependence on the fault mathematical models of power electronics circuit, because the fault mathematical model is difficult to establish.

The fault diagnosis classifiers trained by the time series, original transient features and transient synthetic features are compared in this paper, the classifier obtained a diagnosis accuracy of 96.66% with original transient features. The classifier trained by the transient synthetic features obtained more than 1% gain in performance compared with the original transient features, and is slightly better than time series. The transient synthetic features can be used to improve the classification ability of the fault diagnosis classifier.

Finally, the time scale is considered in the online fault diagnosis system since most AI(artificial intelligence) algorithms are more suitable for running on computer or micro industrial computer. The bottom controller can send 200 sets of three-phase AC currents transient fault data per 20 ms to the computer through the network, and the final fault locations are achieved by using 200 diagnosis results. The online fault diagnosis experiments are presented to prove the effectiveness of the proposed method, the OCFs diagnosis results show that the method can accurately locate the OCFs, and meanwhile the proposed method has a strong universality.

## Declaration of competing interest

The authors declare that they have no known competing financial interests or personal relationships that could have appeared to influence the work reported in this paper.